\documentclass[10pt,twocolumn,letterpaper]{article}

\usepackage{wacv}              % To produce the CAMERA-READY version
\usepackage{graphicx}
\usepackage{amsmath}
\usepackage{amssymb}
\usepackage{booktabs}
\usepackage{comment}

\usepackage[pagebackref,breaklinks,colorlinks]{hyperref}

\newcommand\blfootnote[1]{%
  \begingroup
  \renewcommand\thefootnote{}\footnote{#1}%
  \addtocounter{footnote}{-1}%
  \endgroup
}

\usepackage[capitalize]{cleveref}
\crefname{section}{Sec.}{Secs.}
\Crefname{section}{Section}{Sections}
\Crefname{table}{Table}{Tables}
\crefname{table}{Tab.}{Tabs.}

\def\wacvPaperID{2149} % *** Enter the WACV Paper ID here
\def\confName{WACV}
\def\confYear{2025}

\begin{document}

\title{Agtech Framework for Cranberry-Ripening Analysis Using \\ Vision Foundation Models}

% \title{Agtech Framework for Cranberry Bog Time Series using \\ Vision Foundation Models}

% \title{Vision Foundation Models and an Agtech Framework for Cranberry-Ripening Analysis}

% \title{Vision Foundation Models for Agtech: A Cranberry Bog Time Series Analysis}

\author{Faith Johnson$^{1,*}$
% {\small $^1$ECE Department, Rutgers University}\\
\and
Ryan Meegan$^{1}$
\and
Jack Lowry$^{1}$
\and
Peter Oudemans$^{2,3}$
\and
Kristin Dana$^{1}$
% \vspace{10pt}
% \and
% {\small $^1$ECE Department, Rutgers University ~~~~~ $^2$Department of Plant Biology, Rutgers University}
% \and
% {\small $^3$Philip E. Marucci Center for Blueberry and Cranberry Research and Extension, Rutgers University}
}

\def\keyFont{\fontsize{8}{11}\helveticabold }
% \def\firstAuthorLast{Johnson {et~al.}} %use et al only if is more than 1 author
% \def\Authors{Faith Johnson\,$^{1,*}$, Jack Lowry\,$^{1}$, Kristin Dana\,$^{1}$ and Peter Oudemans\,$^{2,3}$}
% Affiliations should be keyed to the author's name with superscript numbers and be listed as follows: Laboratory, Institute, Department, Organization, City, State abbreviation (USA, Canada, Australia), and Country (without detailed address information such as city zip codes or street names).
% If one of the authors has a change of address, list the new address below the correspondence details using a superscript symbol and use the same symbol to indicate the author in the author list.
% \def\Address{$^{1}$ ECE Department, Rutgers University, New Brunswick, NJ, USA \\
% $^{2}$ Department of Plant Biology, Rutgers University, New Brunswick, NJ, USA \\
% $^{3}$ Philip E. Marucci Center for Blueberry and Cranberry Research and Extension, Rutgers University, Chatsworth, NJ, USA}
% The Corresponding Author should be marked with an asterisk
% Provide the exact contact address (this time including street name and city zip code) and email of the corresponding author
% \def\corrAuthor{Faith Johnson}

% \def\corrEmail{faith.johnson@rutgers.edu}

\maketitle

%%%%%%%%% ABSTRACT
\begin{abstract}
   Agricultural domains are being transformed by recent advances in  AI and computer vision that support quantitative visual evaluation. Using aerial and ground imaging over a time series, we develop a framework for characterizing the ripening process of cranberry crops, a crucial component for precision agriculture tasks such as comparing  crop breeds (high-throughput phenotyping) and detecting disease.
   % as ripening outliers.
  Using drone imaging, we capture images from 20 waypoints  across multiple bogs, and using ground-based imaging (hand-held camera), we image same bog patch using fixed fiducial markers. Both imaging methods are repeated to gather a multi-week time series spanning the entire  growing season.  % Photometric calibration is done for  albedo recovery and berry segmentation is accomplished with recent fine-tuned models and foundation models. 
 Aerial imaging  provides multiple samples to compute a distribution of albedo values. 
 %to assess ripening rate, 
 Ground imaging enables tracking of individual berries for a detailed view of berry appearance changes.   
  Using vision transformers (ViT) for feature detection after segmentation, we extract a high dimensional feature descriptor of berry appearance. Interpretability of appearance is critical for plant biologists and cranberry growers to support crop breeding decisions (e.g.\ comparison of berry varieties from breeding programs).   For interpretability, we create a 2D manifold of cranberry appearance by using a UMAP dimensionality reduction on ViT features. This projection  enables  quantification of ripening paths and a useful metric of ripening rate. We demonstrate the comparison of four cranberry varieties based on our ripening assessments.  
  %Drone imaging samples the crop appearance, observing spatial locations that are similar but not exactly localized and
  %from pixels, and berry segmentation with semi-supervised deep learning networks using point-click annotations.
   This work is the first of its kind and  has future impact  for cranberries and for other crops including wine grapes, olives, blueberries, and maize.  Aerial and ground datasets are made publicly available. 
\end{abstract}

%-------------------------------------------------------------------------
% \subsection{Cross-references}

% For the benefit of author(s) and readers, please use the
% {\small\begin{verbatim}
%   \cref{...}
% \end{verbatim}}  command for cross-referencing to figures, tables, equations, or sections.
% This will automatically insert the appropriate label alongside the cross-reference as in this example:
% \begin{quotation}
%   To see how our method outperforms previous work, please see \cref{fig:onecol} and \cref{tab:example}.
%   It is also possible to refer to multiple targets as once, \eg~to \cref{fig:onecol,fig:short-a}.
%   You may also return to \cref{sec:formatting} or look at \cref{eq:also-important}.
% \end{quotation}
% If you do not wish to abbreviate the label, for example at the beginning of the sentence, you can use the
% {\small\begin{verbatim}
%   \Cref{...}
% \end{verbatim}}
% command. Here is an example:
% \begin{quotation}
%   \Cref{fig:onecol} is also quite important.
% \end{quotation}

% Please number all of your sections and displayed equations as in these examples:
% \begin{equation}
%   E = m\cdot c^2
%   \label{eq:important}
% \end{equation}
% and
% \begin{equation}
%   v = a\cdot t.
%   \label{eq:also-important}
% \end{equation}

%%%%%%%%% BODY TEXT
\vspace{-5pt}
\section{Introduction}
\label{sec:intro}
Machine learning and computer vision methods  play an increasingly vital role in facilitating agricultural advancement by giving real time, actionable crop feedback \cite{luo2023semantic,yin2022computer,meshram2021machine}.  
\blfootnote{\hspace{-15pt}$^*$ Corresponding author: faith.johnson@rutgers.edu\\ $^1$ECE Department, Rutgers University\\ $^2$Department of Plant Biology, Rutgers University\\ $^3$ Philip E. Marucci Center for Blueberry and Cranberry Research and Extension, Rutgers University}
These methods are enabling farming practices  to adapt and evolve to keep up with changing conditions. 
Cranberry farmers are particularly poised to benefit from vision-based crop monitoring 
as they face
numerous challenges related to fruit quality such as fruit rot and over heating \cite{oudemans1998cranberry, polashock2009north, vorsa2012american, vorsa2019domestication}. 
As cranberries ripen and turn red, they become much more susceptible to overheating, partially because they lose their capacity for evaporative cooling \cite{kerry2017investigating,racsko2012sunburn,smart1976solar}. When this growth stage is reached, 
the cranberries exposed to direct sunlight can % overheat and 
become unusable. 
%if not properly watered. 

\begin{figure}
    \centering
    \includegraphics[width=0.23\textwidth]{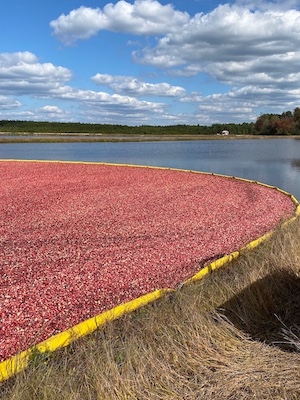}
    \includegraphics[width=0.225\textwidth]{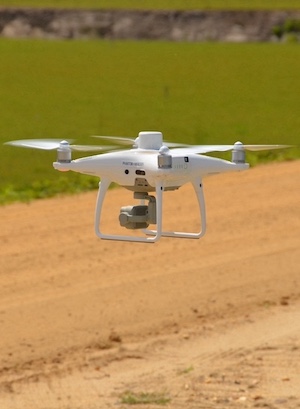}
    \caption{Cranberry bog at the measurement site. (Left) Cranberry harvesting. (Right) Drone at bog for in-field cranberry measurements during the growing season. 
    % put in location after paper acceptance
    }
    \label{fig:teaser}
    \vspace{-15pt}
\end{figure}

We develop a  vision-based method for measuring in-field cranberry albedo to quantify ripening in order to predict when cranberries are nearing this vulnerable stage. 
Currently, cranberry growers quantify this ripening process manually using out-of-field  albedo evaluation by imaging harvested cranberries over time \cite{oceanSprayPerCom}.
%to track when their crops are nearing this vulnerable stage
%While this method may be accurate,
%While the out-of-field albedo evaluation may be accurate, it does not allow for expedient in-field decision making, meaning that by the time farmers find out their crop is at risk, it may be too late to intervene. 
% I commented out the above line because it gave the impression that the growers need to pick and measure the cranberries to determine if they are red enough to warrant watering
This approach is cumbersome and time-consuming, limiting its utility in larger-scale evaluations. For practical applications, only small numbers of berries can be harvested for out-of-field images.  
%These irrigation decisions are carefully planned because unnecessary irrigation leads to increase cost and increased risk of fungal rot. 
%In-field imaging has the advantage of showing a larger sampling of berries and seeing potential differences between bogs of the same variety or between different spatial regions of the bog.
%
This is inefficient and incapable of painting a full picture of overall crop health as plants do not ripen uniformly. Berries on the top of the canopy with direct sun exposure have a high risk of overheating, while berries underneath the leafy canopy are generally well protected.

\begin{figure}
    \vspace{-5pt}
    \centering
   \includegraphics[width=0.45\textwidth]{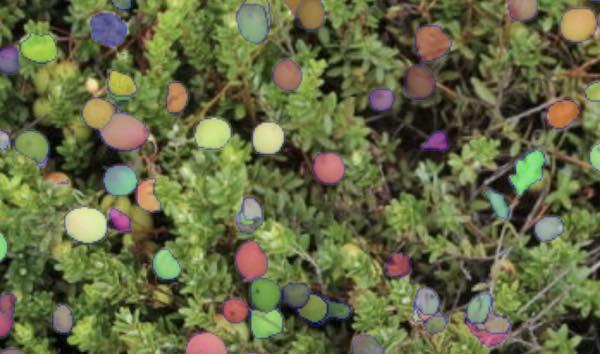}
   \caption{An example segmentation of cranberry images using the Segment Anything Model SAM\cite{ravi2024sam} without point-click prompts (automatic mask generation). Notice that in addition to cranberries, surrounding leaves and other structures are segmented.}
   \label{fig:exampleSAM}
   \vspace{-5pt}
\end{figure}

The conventional solution to overheating is increased crop irrigation during the growing season. 
%Inadequate watering speeds up crop rot due to overheating, but over-watering also leads to mass cranberry crop loss due to an increased risk of fungal infection 
These irrigation decisions must consider cost and efficient use of environmental resources. Furthermore, inadequate or poorly timed irrigation can lead to overheating whereas excessive irrigation encourages fungal fruit rot to develop
\cite{oudemans1998cranberry}. 
Since berries can overheat in a short period of time \cite{pelletier2016reducing,kerry2017investigating}, irrigation decisions should be coordinated with in-field albedo characterization, informing the grower on the number of vulnerable berries and enabling expedient decision making.
%There are cost considerations for not over-watering a crop and maximizing the efficient use of environmental resources. 
%Additionally, only those berries in the top of the canopy need to be watered, as they have the most sun exposure and the highest risk of overheating. 
 Therefore, assessing the current albedo of the visible berries is directly relevant to irrigation decisions.
%Knowing the expected ripening rate for a particular cranberry bog 
For these reasons,  in-field measurement of the ripening rate for a particular cranberry bed
significantly informs crop management decisions. 

%Computer vision segmentation provides a real time  tool to assist growers to assess irrigation issues.

For in-field measurements, we use drone imaging (see Figure~\ref{fig:teaser} and ground-based imaging with standard RGB cameras (digital SLR). 
Our ripening assessment framework uses cranberry image segmentation to evaluate albedo variation over time to compare cranberry varieties. We conduct this albedo analysis on two spatial scales: broad area (bog-wide) and on the individual berry level.
Ripening rates vary among cranberry varieties, and ones that ripen early are at the greatest risk. 
While surveying ripening rates over an entire bog can provide ripening statistics over the group, the appearance changes of an individual berry allows for a more detailed assessment of traits and variations of the berry ripening patterns. 
%Solely surveying ripening rates over an entire bog, where multiple varieties can coexist during the growing season, may lead to an incomplete ripening assessment. 

In prior work \cite{akiva2022vision,akiva2020finding},  neural networks for segmentation have been used for yield estimation through counting. In our work, we use these cranberry-tuned segmentation networks for bog-wide albedo analysis. For the individual berry analysis, we use recent foundation model segmentation networks (SAM \cite{ravi2024sam}) to isolate individual berries over time to find temporal patterns in cranberry albedo across varieties. The foundation models are useful when point-click prompts can be given for individual berries, while the cranberry-tuned methods are useful for large areas/bogs since they require no point-click prompts.  Without point-click prompts, SAM-based segmentation segments both cranberries and leaves as shown in Figure~\ref{fig:exampleSAM}.

\begin{figure}
    \vspace{-5pt}
    \centering
   \includegraphics[width=0.45\textwidth]{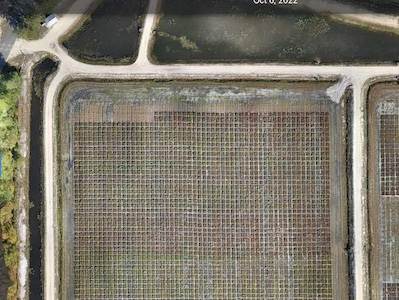}
    \caption{Example of breeding plots (drone view) that are typically evaluated manually.  Planting design permits approx 3500 plots/ha, and this entire block is approximately 2 ha.  Convenient quantitative evaluation can be supported by our vision-based ripening assessment framework. {\it Location removed for blind review.}}
    \label{fig:HTP}
    \vspace{-5pt}
\end{figure}
\begin{comment}
By combining counting and albedo analysis, 
it becomes possible to evaluate the economic risk to a particular crop on a high temperature day, while also making a long term assessment on which varieties provide the most yield.

The cranberries are segmented using a  semi-supervised method \cite{akiva2022vision} based on the Triple-S network \cite{akiva2020finding}. Using point-wise annotations instead of pixel-wise labels significantly reduces the labeling cost for the densely populated cranberry images. 
We provide new point-click labeled cranberry imagery that supports ripening assessments in CRAID4, a new dataset covering a larger time period of the growing season with more temporal frequency than prior work.
We train a segmentation network to isolate cranberry pixels from drone imagery in both this new dataset and the pre-existing CRAID1 dataset \cite{akiva2020finding}. 
\end{comment}
Using the framework of imaging, photometric calibration, and cranberry segmentation, and albedo analysis for estimation a ripening metric: we  present a  ripening comparison of four cranberry varieties over a two month span and show clear timelines of albedo change indicating when each variety becomes at great risk for overheating.

\subsection{Impact for Crop Breeding} Screening for the heritability of novel genotypes requires high through-put phenotyping (HTP) methods to discover desirable genetic traits \cite{araus2014field, diaz2018massive}. In crop breeding, there may be hundreds to thousands of progeny/offspring to evaluate, and HTP methods make this evaluation practical. Computer vision algorithms for segmentation and calibrated albedo measurements enable quantitative comparisons.  The  methodology we present in this paper fits those requirements well, and HTP is an application domain for this work.  

\begin{figure*} 
    \vspace{-10pt}
    \includegraphics[width=\textwidth]{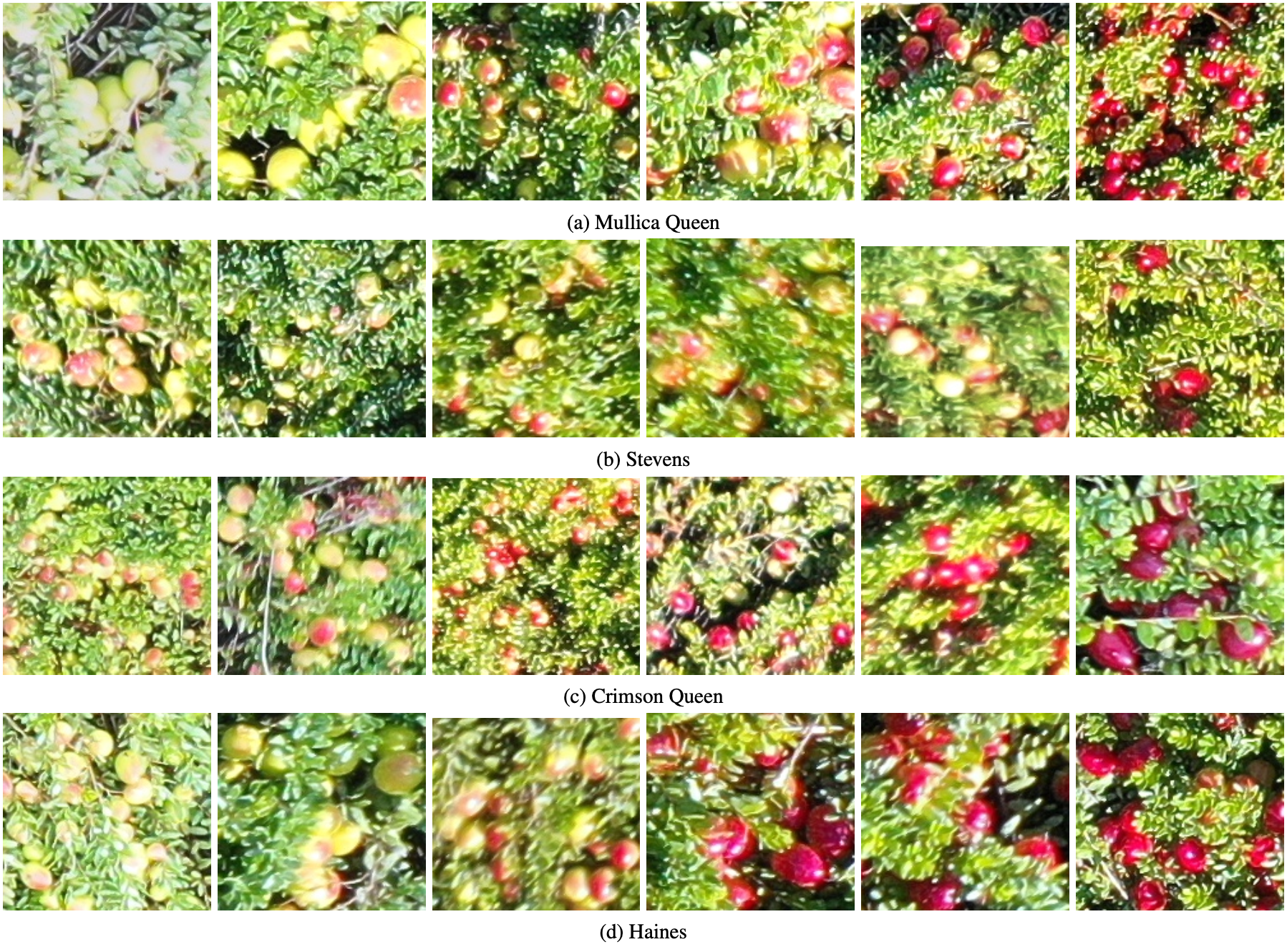}
    \vspace{-20pt}
    \caption{Drone images from multi-temporal drone-scout imaging. Weekly inspection of multiple cranberry bogs over the late July/September growing season for four varieties (Mullica Queen, Stevens, Crimson Queen, Haines). (Left to Right) Imaging Dates for 2022: 7/27, 8/2, 8/16, 8/25, 8/31, 9/9.  }

    \label{fig:cranOverTime}
    \vspace{-5pt}
\end{figure*}

The rate of color development is a crop trait that can affect the quality of cranberries at harvest. For consumer appeal,  the timing and uniformity of ripening is critical, i.e. asynchronous ripening is a problem. For breeding, uniformity is desirable so   HTP is used to look at multiple genotypes. For example, our related current work (unpublished) evaluates 300-400 genotypes planted in small plots (e.g. 3.3 sq.\textbackslash m.) where a ripening evaluation is done out-of-field and only a few times (1-2) per season depending on time and labor.  To illustrate the scale of these studies, consider that they include 0.5 acre plots  with 350 individual small plots and 
0.2 hectare (ha) plots with 7000 individual plots (see drone image shown in Figure~\ref{fig:HTP}).

\section{Related Work}

% and highlight that those were about counting and this is about learning the temporal signatures of the crop(s)

% Temporal Albedo Sequence for Crop Comparison --> $TASC^2$?
% Temporal Albedo Characterization? --> TAC?

\subsection{Precision Agriculture}

Precision agriculture is revolutionizing farming and challenging traditional methods. Future farms will integrate multiple technical advances such as soil sensors \cite{abdollahi2021wireless}, plant wearables \cite{yin2021soil}, drone aerodynamics \cite{radoglou2020compilation}, and remote sensing \cite{sishodia2020applications}.
Advances in machine learning and AI have been particularly impactful, enabling significant breakthroughs in agricultural applications in recent years \cite{wang2022review,benos2021machine,sharma2020machine,mavridou2019machine}. Computer vision is capable of giving real time, high fidelity feedback to farmers about yield estimation \cite{van2020crop,darwin2021recognition,he2022fruit,palacios2023early}, phenotype identification \cite{li2020review,kolhar2023plant,liu2022tomatodet}, and crop health assessment \cite{dhaka2021survey,ahmad2022survey,kattenborn2021review} while also being useful for larger scale applications like farm automation \cite{friha2021internet}. 

\subsection{Albedo Characterization over Time}
% \begin{itemize}
    % \item cranberry spoilage risk goes up as they ripen
    % \item bc can't do evaporative cooling
    % \item detecting when this change occurs can help farmers better utilize their resources to prevent crop rot
    % \item this ripening visually corresponds to a change in albedo 
    % \item we aim to measure this change over time with images in order to characterize the temporal albedo signature of different cranberry varieties
% \end{itemize}

 As cranberries ripen, their risk of spoilage increases due to overheating \cite{pelletier2016reducing} caused by a decrease in evapotranspiration. 
  This ripening corresponds with visual changes in the berry albedo, which allows us to use albedo characterization over time to predict when a cranberry bog is most at risk.
 This same phenomena is also found in apples \cite{racsko2012sunburn} and grapes \cite{smart1976solar}.
 %,gambetta2021sunburn}. 
 Ripening patterns can also indicate the presence of viruses as occurs in wine grapes \cite{alabi2016impacts,blanco2017red}.
  Despite the importance of quantifying color development, automated methods for albedo characterization have received limited attention in the literature. 
 Most existing studies of ripening, do out-of-field measurements that rely on harvested berries for evaluating ripening \cite{vorsa2017performance, keller2010managing}. 
 These methods  are time-consuming and do not scale to large evaluations or real-time assessments. 
 The framework of this paper is an important step for using computer vision methods (foundation models,  deep learning networks, and classic vision methods) as a tool in agriculture. 

 %We segment the cranberries periodically over the growing season and pinpoint when each variety is most likely to overheat. Detecting when this change occurs can help farmers better utilize their resources to prevent crop rot. 

% for reference, Peri's journal paper references the different craids by year so there's Craid-2019, Craid-2021, and now we're releasing Craid-2022
\begin{figure}
    \vspace{-5pt}
    \centering
    \includegraphics[width=\linewidth]{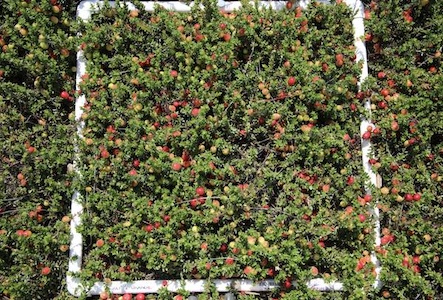}
    \caption{Ground-based imaging (hand-held DSLR camera) of the same region in a cranberry bog was done over 27 sessions (almost daily during the growing season) to show the appearance of individual cranberries over time. The semi-permanent PVC frame enabled identification of the imaged region over time.   }
    \label{fig:pvc}
    \vspace{-10pt}
\end{figure}

\section{Methods}  
\subsection{Cranberry Bog Drone Imaging}

We introduce CRAID-4, a new dataset from bog monitoring with drones.
%%% UNCOMMENT the below line for the CAMERA READY
%at PE Marucci Center for Blueberry and Cranberry Research, a substation of the Rutgers New Jersey Agricultural Experiment Station (Chatsworth, NJ). 
We combine the  CRAID-1 dataset \cite{akiva2020finding} with our new drone-based cranberry bog images that we call CRAID-4.
%%% REMOVE THIS FOR CAMERA READY
%from the 2022 growing season. 
In this work, the term CRAID$+$ dataset refers to the combination of CRAID-1 and CRAID-4.
%taken at a farm in Chatsworth, NJ.
In total, this dataset contains four different cranberry varieties over seven bogs. The cranberry variety names are {\it Mullica Queen, Stevens, Haines,} and {\it Crimson Queen}.
The images include three beds of Mullica Queen, one bed of Stevens, two beds of Haines, and one bed of Crimson Queen cranberries, as shown in Figure~\ref{fig:cranOverTime}. The images were taken by drone in weekly increments between the months of July and September. 
% to capture the full range of cranberry appearances during the growing season.
% 3412 crops total used for the albedo analysis
% craid 4 has 82296 crops total
% take 72 crops per image (8 by 9)
% labeled 2 additional images from each day on each bog for a total of 220
% craid1 is 2800 total labeled images with 2368 for training
We calibrate each drone image and crop each into 72, non-overlapping $456 \times 608$ sub-images used for training the cranberry segmentation network \cite{akiva2022vision}. A selection of 220 crops representative of the diverse berry appearances in the entire growing season were manually labelled with point-wise annotations for all berries in the image. This data was combined with the labeled dataset of 2368 images from \cite{akiva2020finding} to create a new training dataset comprised of 2588 total images.
%resulting in a dataset comprised of 3 seasons of cranberry data. 
We train on this combined dataset of 2588 images (CRAID$+$). 
%as well as the smaller dataset of 220 images (CRAID-4 only)
%and compare the pipeline performance between the two.
The resulting segmentation is high quality (visually assessed) and requires no point-clicks after training.

% \begin{itemize}
    % \item CRAID4 is naturally more skewed towards green than red berries as opposed to CRAID1 because it's taken across a larger portion of the growing season than CRAID4 (which was mostly taken at the end??? I don't actually know if this sentence is true) 
    % \item We chose a selection of 220 cropped images that represented a the entire growing season and berry appearances, and manually labelled pointwise pointwise annotations for all berries in the image.
    % \item This data was combined with the labelled dataset of 2368 images from \cite{akiva2020finding} to create a new training dataset.
% \end{itemize}

% \subsection{Cranberry Imaging}
% \begin{itemize}
%     \item{drone used - dji mavic 3(?)}
%     \item{which year - } 
%     \item can we just say we followed the same procedure as \cite{akiva2022vision}????
%     % \item{Prior work provided results and annotation for Craid 1 and 2?, In this work we move to craid 4, provide new annotations and demonstrate transfer learning from prior algorithms to the new set}. 
% \end{itemize}

%%%
% In the photometric calibration part, the description of the calibration method in the article is very unclear. For example, the changes before and after calibration are not mentioned, and the formula of relevant transformation is not described
%%%

\subsection{Individual Cranberry Imaging}
\label{sec:individualImaging}
For single berry tracking, our goal is to image {\it the same berry} over a time sequence to characterize ripening and appearance in a more precise manner.  To this end, in addition to drone imaging,  we manually captured the same $12\times 12$ region of the cranberry bog for 27 days sampled between August 7th, 2023 to September 22nd, 2023.
% almost daily (27 days sampled  from  ).  
A semi-permanent fiducial marker was constructed as a square of PVC pipes (see Figure~\ref{fig:pvc}) to denote the area for image capture. The images (of size 1361x907) of this region over time are registered using SIFT \cite{lowe2004distinctive}, specifically the FLANN based feature matcher with 20 trees and 200 checks  \cite{muja2009fast}. We align each image in this time series to the first image of the sequence using a homography estimated with RANSAC \cite{fischler1981random}.  To derive the cranberry segmentations, we use SAM 2 \cite{ravi2024sam} where 
%specifically the \verbatim{sam2_hiera_large model}.
 point clicks are passed into this SAM model’s image predictor class.   These point clicks are necessary for this task, instead of using the automatic mask generation, to avoid segmentation of crop leaves and shaded regions (see Figure~\ref{fig:exampleSAM}). 
Using this approach we obtain the first time-lapse cranberry imaging series, key for computing ripening metrics for crop assessment and variety comparisons.

\begin{figure}
    \vspace{-5pt}
    \centering
    \includegraphics[width=0.33\linewidth]{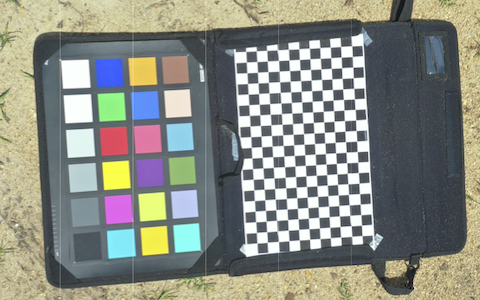}
    \includegraphics[width=0.32\linewidth]{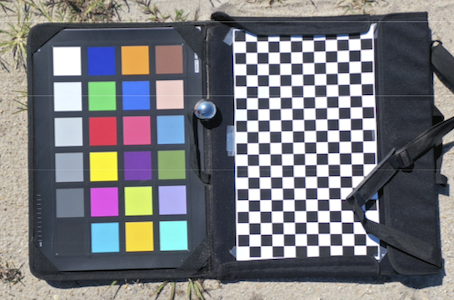}
    \includegraphics[width=0.33\linewidth]{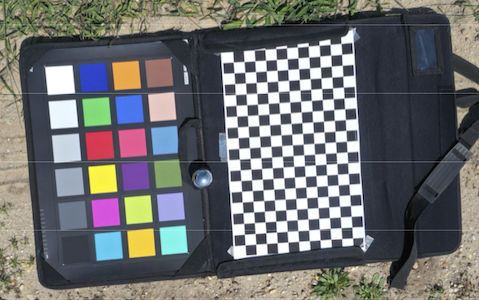}
    \caption{Images for photometric calibration. The drone imaging protocol includes images of  the Macbeth card over multiple days. Although the same camera is used for imaging, camera parameters change. Notice slight variations in card appearance from different days, which we remove through photometric calibration.}
    \label{fig:macbeth}
    \vspace{-5pt}
\end{figure}
\subsection{Photometric Calibration}
The images in CRAID-4 %from the 2022 growing season 
were first photometrically calibrated using the Macbeth Color Checker card (shown in Figure~\ref{fig:macbeth}) and a well-established approach of estimating the optimal radiometric correction using measurements of the card under uniform illumination \cite{kim2008robust, debevec2008recovering,mitsunaga1999radiometric}. 
Radiometric or photometric calibration is needed to account for the effects of the changing camera parameters and sun angle between imaging sessions.
% and the appearance variations caused by thechange in sun angle. 
For an invariant albedo measurement,  raw pixel values from images are insufficient since they depend on camera parameters and environment conditions. 
Reference images of the card were taken from every bog for each day of data collection using the drone camera. For each reference image we extracted intensity values for the 6 grey scale squares on the Macbeth Color Checker. The measured values were used to find a linear transformation to recover the radiometric correction parameters, and the images were corrected accordingly. 

\subsection{Bog Albedo Analysis}
The industry standard for ripeness defines five classes of cranberries based on albedo \cite{oceanSprayPerCom}. % change over time. 
These distinct stages of ripeness are hand-defined by field experts from the periodic collections of the cranberries throughout the season \cite{oceanSprayPerCom}. We use a similar approach for classifying albedo change in the CRAID data, but opt for using in-the-field images of the cranberries in the bog instead. 
%Each class is defined by TSNE clustering the RGB pixel values of a collection of randomly sampled cranberry detections from the entirety of the images from the 2022 growing season. We cluster points in this embedding space with k-means,
In the visual analysis of cranberry images, each class is defined by k-means clustering  the RGB pixel values of a collection of randomly sampled cranberry detections sampled over the entirety of the  growing season (k=5). We then map those clusters to the 5 ``common classes", spanning from green to red, that best align with the industry standard. This human-in-the-loop mapping combines  domain knowledge of the industry-standard classes with the automated clustering results. 

\begin{figure*}
    \centering

    \vspace{-5pt}
    \includegraphics[width=\textwidth]{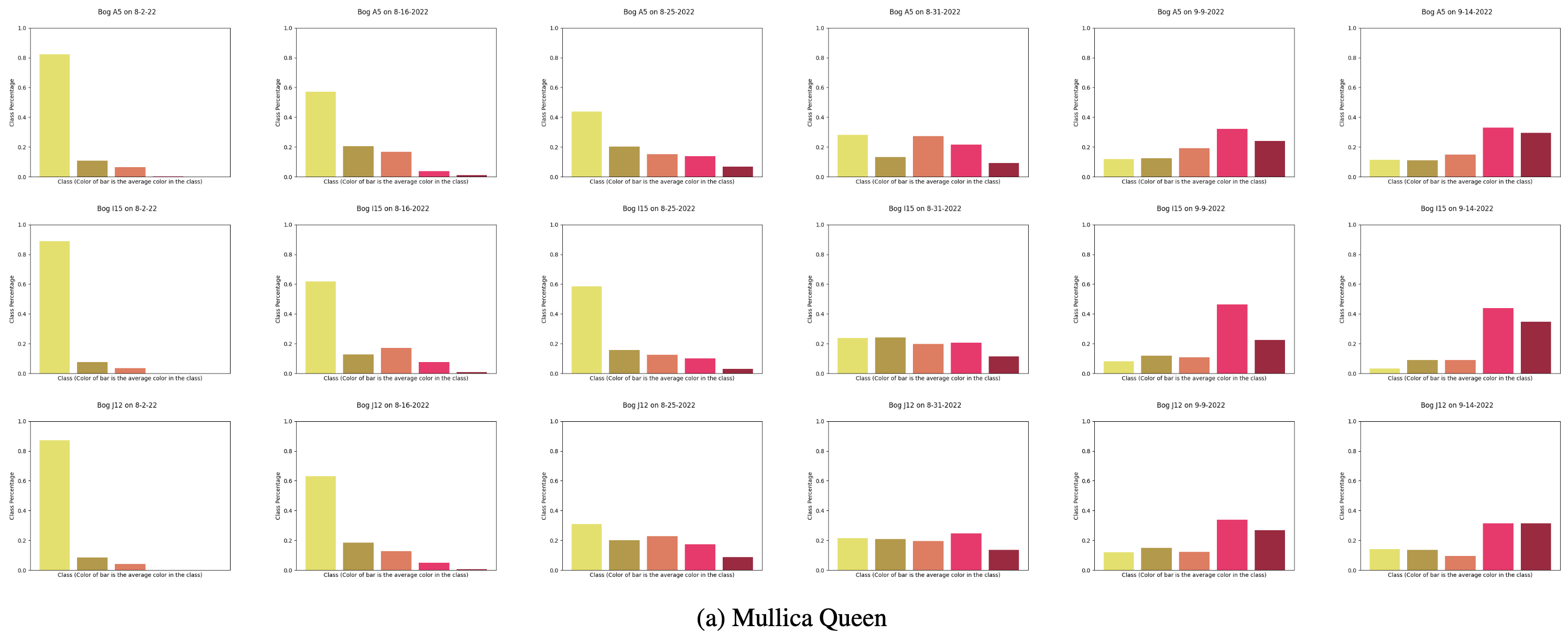}
    \includegraphics[width=\textwidth]{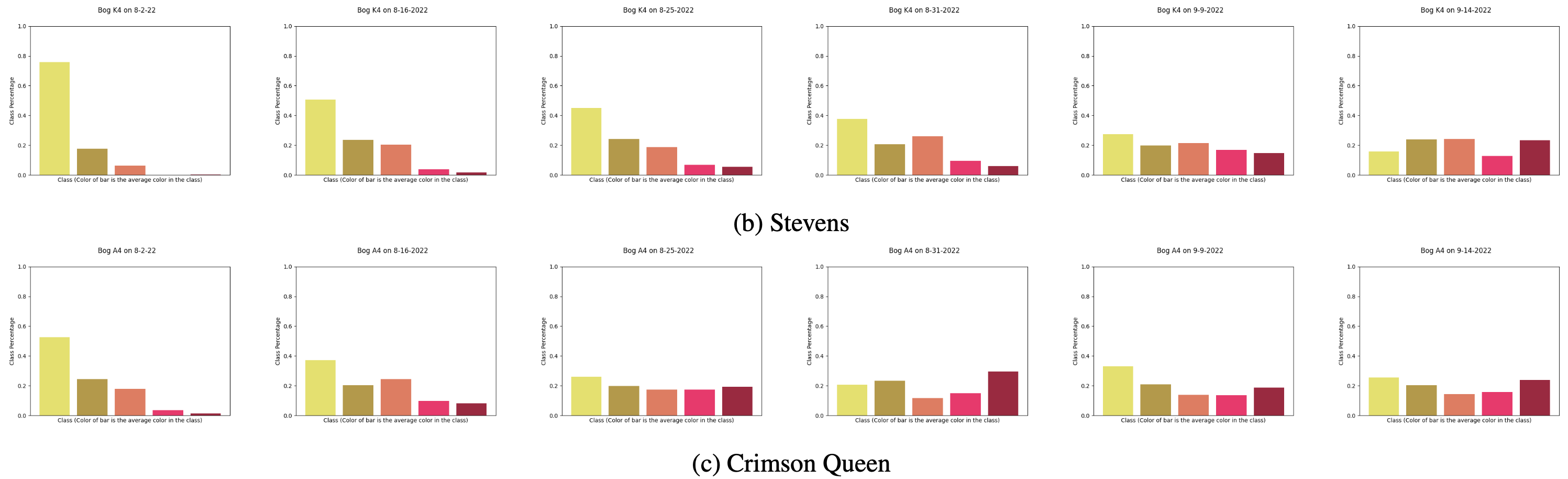}
    \includegraphics[width=\textwidth]{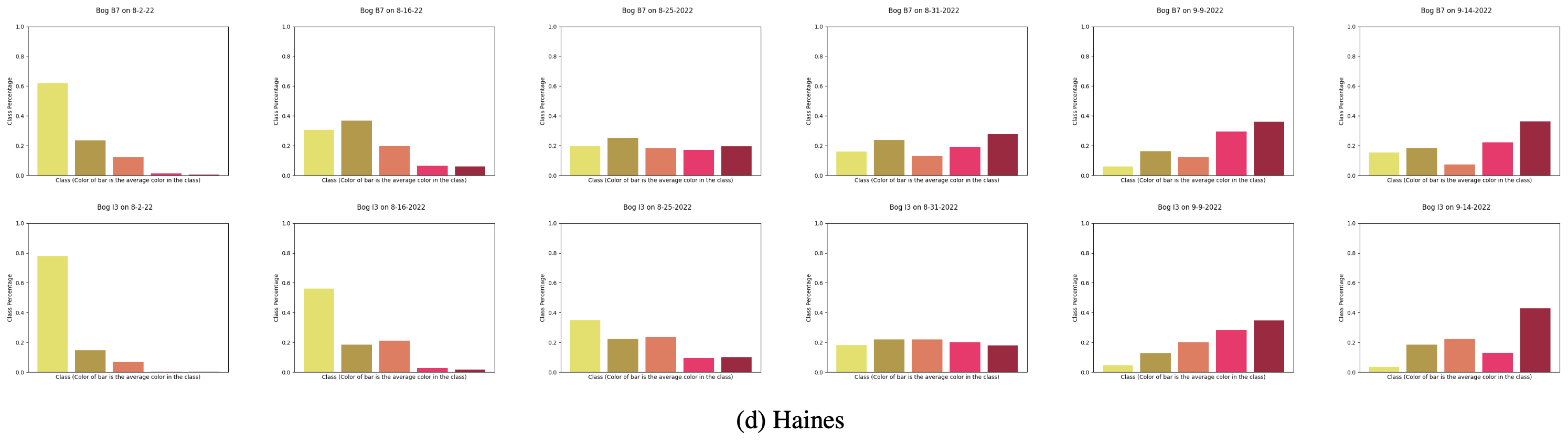}
    \vspace{-15pt}
    \caption{Plots comparing albedo over time for four cranberry varieties. Histograms of pixels in the five main color classes are shown. The four varieties are: Mullica Queen (top three rows), Stevens, Crimson Queen, and Haines (bottom two rows). Residual green pixels at the later dates are artifacts due to some misclassifications of background leaf pixels. }
    \label{fig:albedoOverTime}
    \vspace{-5pt}
\end{figure*}

Once the classes have been determined, we match each berry to its corresponding color class by matching each pixel belonging to a single berry to its closest color cluster. The cluster belonging to the majority of the pixel values is chosen as the label for the berry. We repeat this process for each cranberry and count the number of detections in each image. The change in class density is plotted, as in Figure~\ref{fig:albedoOverTime}, and clearly shows patterns in the cranberry albedo. From the progression of these plots, we also pinpoint when each cranberry variety becomes most at risk of overheating. 

% \begin{itemize}
    % \item The industry has a way of evaluating the appearance. We can match a clustering to that description as follows:  We cluster with k-means then map those clusters to the 5 "common classes"
    % \item{We need 3D rgb plots}
    % \item{We need to tell the story of these rgb plots}
    % \item{Need plots as in Peters 5 class plots}
    % \item We extracted the albedo values from each segmented cranberry pixel, and ran K-Means to extract 11 albedo clusters
    % \item take a random sample of berries, tsne the rgb points, then kmeans the tsne embeddings,
% \end{itemize}

\subsection{Individual Berry Albedo Analysis}
\label{sec:individualBerryAnalysis}

\begin{figure*}
  \centering 
    \includegraphics[width=3.55in]{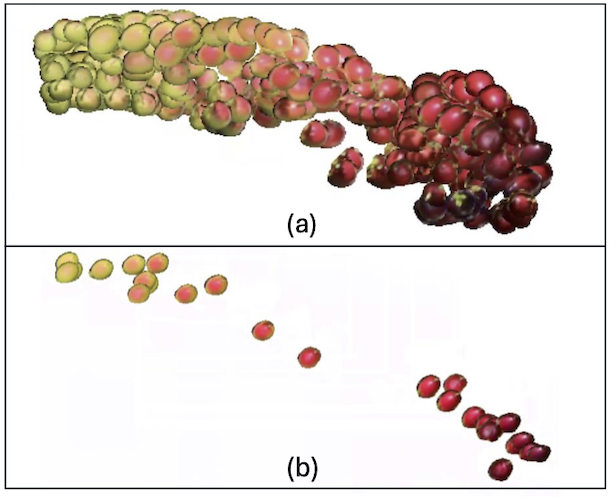}
       \caption{(a) Segmented cranberry images projected to a 2D manifold using UMAP on ViT features.  Berry images are shown in the location of the projected UMAP features. (b) An individual berry tracked over 27 timepoints (from 6 weeks) and projected to the learned 2D appearance manifold revealing its ripening path.  }
    
    \label{fig:cranvitsingle}
    \vspace{-5pt}
\end{figure*}

Fourteen individual berries from one bog were tracked over a time series consisting of  27 time points (over 6 weeks). Once the individual berries were tracked using alignment and segmentation, as described in Section \ref{sec:individualImaging}, visual features were extracted from each individual berry image. We employ and compare the feature vector quality of four different off-the-shelf feature extractors: DinoV2 Giant \cite{oquab2023dinov2}, Google ViT (Vision Tranformer) Huge \cite{dosovitskiy2020image}, SAM 2 Hiera Huge \cite{ravi2024sam}, and Laion CLIP Big G \footnote{implementations from https://huggingface.co/} \cite{schuhmann2022laion}. The resulting features vectors are projected to a 2D manifold using UMAP \cite{mcinnes2018umap} (as shown in Figure~\ref{fig:cranvit}) with the cranberry images rendered at their UMAP locations.  Figure~\ref{fig:cranvitsingle} shows the ViT features projected in UMAP since this feature is selected for the ripeness metric, and an individual berry's trajectory over time is shown.    %in Figures~\ref{fig:cranvit} and ~\ref{fig:cranvitsingle} 
This projection provides an interpretable representation useful for growers and plant biologists. Knowing where the crop currently resides in this manifold in real-time allows for expedient decision-making that can increase crop yield and overall health.

% these are the varieties matched with the Bog numbers
% A4 Crimson Queen
% A5 Mullica Queen
% I5 Mullica Queen
% J12 Mullica Queen
% K4 Stevens
% B7 Haines
% I3 Haines

\section{Results}
\label{sec:results}

    % \begin{subfigure}
    % \includegraphics[width=2.5in]{images/craid4_1.png}
    % \includegraphics[width=2.5in]{images/craid4_1(2).png}
    % % \includegraphics[width=0.33\textwidth]{images/craid4_1(3).png}
    % \caption{Segmentation Map from the Network Trained on  Dataset A (best viewed zoomed).}
    % \label{fig:segMapa}
    % \end{subfigure}
    % \begin{subfigure}
    % \includegraphics[width=2.5in]{images/craid4.png}
    % \includegraphics[width=2.5in]{images/craid4(2).png}
    % % \includegraphics[width=0.33\textwidth]{images/craid4(3).png}
    % \caption{Segmentation Map from the Network Trained on Dataset B (best viewed zoomed). Observe high quality segmentation with only 220 images annotated in the training set.  }
    % \label{fig:segMapb}
    % \end{subfigure}
    
\begin{comment}
    
\begin{figure*}
    \centering
    \includegraphics[width=0.9\textwidth]{images/prediction_comparison.png}

    \caption{Example predicted cranberry segmentation maps overlaid on the original input images. (Best viewed zoomed.) The top row was predicted from the model trained on CRAID$+$. The bottom row was predicted by the model trained solely on the 2022 growing season data, CRAID-2022. The first model (first row) predicts significantly smaller cranberry blobs than the second model (second row). Additionally, the model trained on CRAID$+$ struggles to segment greener berries due to a lack of early images with green berries in the CRAID-2019 dataset that forms the majority of CRAID$+$ dataset. }
    \label{fig:segMap}
    \vspace{-10pt}
\end{figure*}
\end{comment}

\begin{table}[]
    \centering
    \begin{tabular}{|c|c|c|c|c|c|c|}
    \hline
        \multicolumn{7}{|c|}{Ripeness Ratio} \\
        \hline 
        Bog  & 8/2 & 8/16 & 8/25 & 8/31 & 9/9 & 9/14 \\
        \hline
        \hline 
         A5 & 0.007 & 0.082 & 0.331 & 0.497 & 0.902 & 1 \\
         I15 & 0.001 & 0.108 & 0.167 & 0.409 & 0.874 & 1 \\
         J12 & 0.002 & 0.088 & 0.419 & 0.609 & 0.968 & 1\\
         K4 & 0.012 & 0.151 & 0.339 & 0.433 & 0.872 & 1 \\
         A4 & 0.127 & 0.453 & 0.926 & 1.118 & 0.808 & 1 \\
         B7 & 0.035 & 0.217 & 0.622 & 0.798 & 1.119 & 1 \\
         I3 & 0.010 & 0.079 & 0.347 & 0.678 & 1.121 & 1 \\
         \hline
    \end{tabular}
    \caption{Ripeness ratio for each bog of cranberries over time. We define the ripeness ratio for a bog of cranberries to be the percentage of red berries at the current time over the percentage of red berries on the final collection date. 
    Bog key indicates the following cranberry types: A5 Mullica Queen, I5 Mullica Queen, J12 Mullica Queen, K4 Stevens,  A4 Crimson Queen, B7 Haines, I3 Haines. 
    }
    \label{tab:ripenessRatio}
    \vspace{-10pt}
\end{table}
\subsection{Broad Area (Bog) Ripening Metrics}

The cranberry segmentation network has a mean intersection over union (mIOU) of 62.54\% and a mean absolute error (MAE) of 13.46 as reported in \cite{akiva2021ai}. %of the cranberry segmentation network in Table \ref{tab:metrics}. 
% We show the results of the cranberry segmentation from the model trained on CRAID$+$ in Figure \ref{fig:segMap}a alongside the results from the model trained on only the 2022 growing season data (CRAID-4) in Figure \ref{fig:segMap}b. 
Training on the larger CRAID$+$ dataset produces smaller predicted cranberry blobs than training on only the 2022 season data in CRAID-4. This may be due to a scale mismatch between the datasets. The CRAID-1 dataset contained images of cranberries taken by drone from a higher elevation than in the CRAID-4 data. Another deficiency of the model trained on the CRAID$+$ data is that it misses detections of greener berries. Because the CRAID-1 dataset contains mostly red berries, it is unable to make the color invariant predictions necessary to accurately segment green berries in the CRAID-4 dataset.
%CRAID data. 
For this reason, we use the segmentation network trained on CRAID-4 for our albedo characterization.

% \subsection{Albedo Analysis}
% \begin{itemize}
    % \item The industry has a way of evaluating the appearance. We can match a clustering to that r4sddescription as follows:  We cluster with k-means then map those clusters to the 5 "common classes"
    % \item{We need 3D rgb plots}
    % \item{We need to tell the story of these rgb plots}
    % \item{Need plots as in Peters 5 class plots}
    % \item take a random sample of berries, tsne the rgb points, then kmeans the tsne embeddings,
% \end{itemize}

% \begin{table}[]
%     \centering
%     \begin{tabular}{|c|c|c|}
%         \hline
%         & mIOU & MAE \\
%         \hline
%         \hline
%          Cranberry Segmentation Network \cite{akiva2021ai}  & 62.54 &  13.46 \\
%         \hline
%     \end{tabular}
%     \caption{We report the mean Intersection Over Union (mIOU) and Mean Absolute Error (MAE) metrics for the method described in \cite{akiva2021ai}. We use this network to segment the cranberries for our albedo characterization. %For MAE, lower numbers indicate better performance. The opposite is true for mIOU. 
%     }
%     \label{tab:metrics}
% \end{table}

\begin{figure*}
    \centering
    \includegraphics[width=\linewidth]{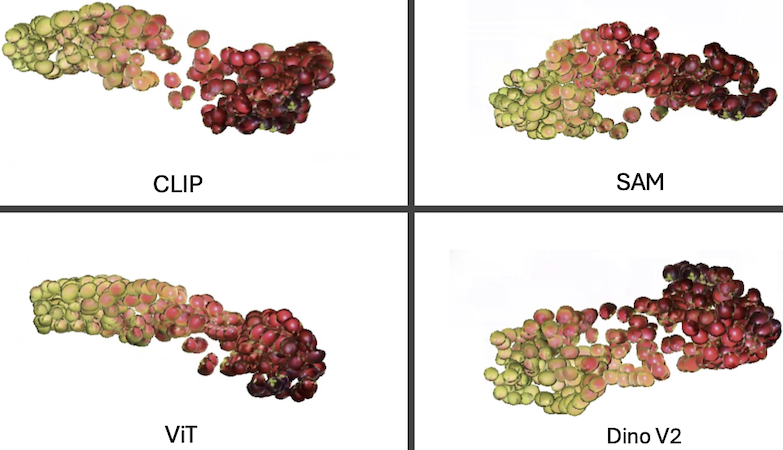}
    \caption{We use four different off-the-shelf feature extractors: Laion CLIP Big G \cite{schuhmann2022laion},  SAM 2 Hiera Huge \cite{ravi2024sam}, Google Vit Huge \cite{dosovitskiy2020image}, and DinoV2 Giant \cite{oquab2023dinov2}.  Fourteen individual berries were imaged for 27 time points and the images were aligned and segmented. Feature extraction of the berry images using these four feature extractors. The resulting features vectors are projected to an interpretable representation using UMAP resulting in the embeddings illustrated here. (All $14\times 27$ berry images are shown here at their corresponding UMAP coordinate).  ViT features led to the most useful 2D embedding, showing ripeness progression in a well-distributed path close to a line.  }
    \label{fig:cranvit}
    \vspace{-5pt}
\end{figure*}

% Figure \ref{fig:rgbAlbedo} illustrates the process by which the cranberries are classified into one of the five common albedo classes. 
Once the berries are successfully segmented, their constituent pixels are matched to the closest color cluster, and each berry is labeled with the cluster that appears most frequently. %Figure \ref{fig:rgbAlbedo} shows the isolation of a cranberry and the distribution of the RGB values belonging to it. This berry is primarily red, so it would be mapped to class 5, which contains the reddest, most ripe berries. 
We compute the classes of all the berries and plot the percentage of berries in each class for a particular collection date and bog in Figure \ref{fig:albedoOverTime}. The top three rows are the Mullica Queen variety. The next row is the Stevens variety followed by the Crimson Queen variety. The final two rows are the Haines variety. Each column of graphs is made from berries imaged on a specific day. From left to right, the columns were imaged on the following dates: 8/2, 8/16, 8/25, 8/31, 9/9, and 9/14; the ripening weeks for cranberry bogs.
% UNCOMMENT the line below stating "in 2022" for CAMERA READY
% in 2022.
As the berries redden, 
 the cranberry bog enters the high risk category and the %there are more berries in the red and dark pink classes than in the green classes. 
the ripeness ratio, as shown in Table \ref{tab:ripenessRatio}, can be used to determine a
ripeness threshold (e.g.  approximately 0.6) as an indicator.  This ripeness ratio is measured as the percentage of red berries (class 4 and 5) on a collection date divided by the percentage of red berries on the final collection date. 

%\textbf{Note from Faith: should this threshold be higher than 0.5? Peter, when would you say that each berry is in danger of overheating based on the graphs?}

\begin{figure*}
\vspace{-5pt}
    \centering
    \includegraphics[width=5.7in]{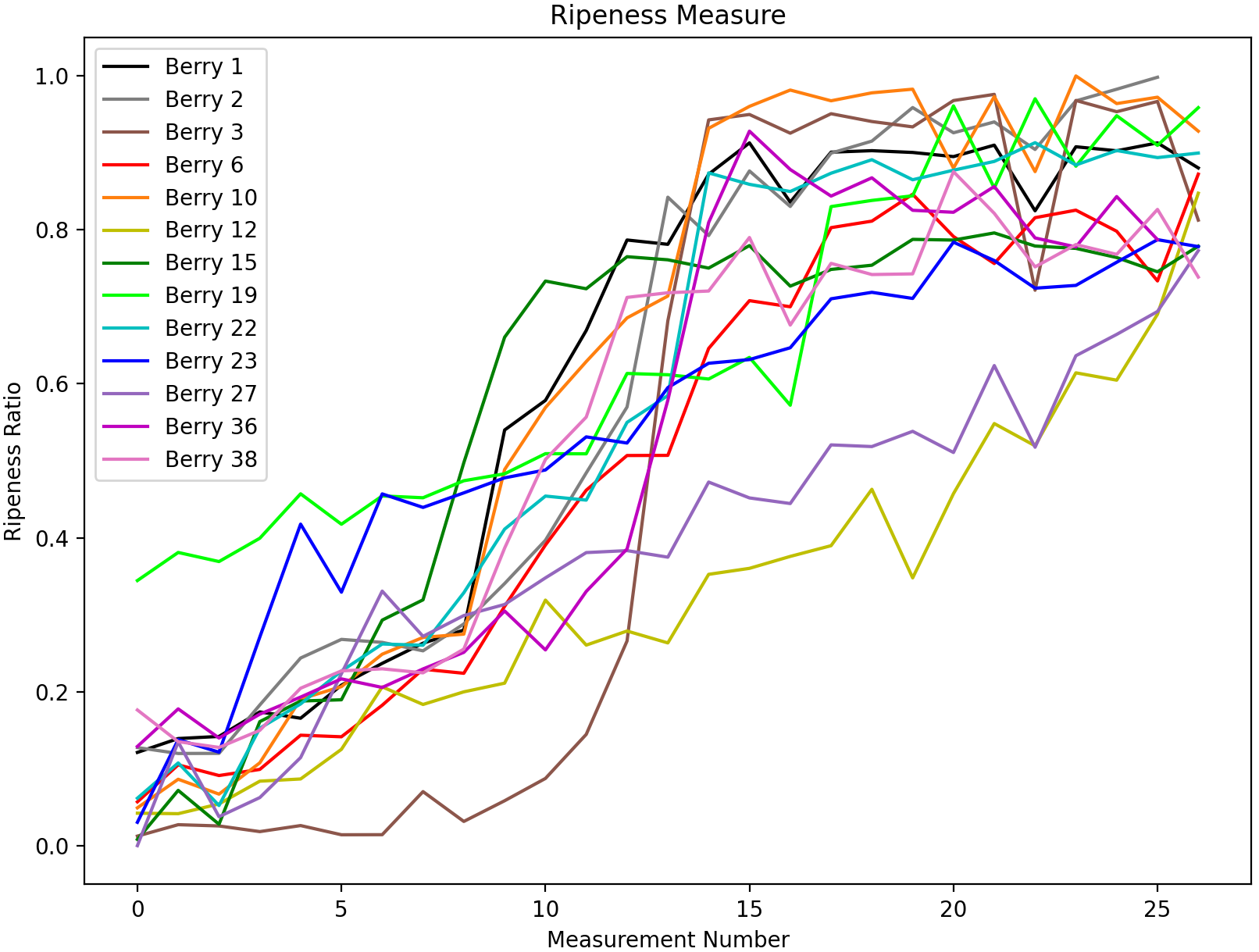}
    \caption{
    Ripeness measure for fourteen individual berries tracked and image over the growing season (27 measurements over 6 weeks). The ripeness ratio is the result of our multi-step framework consisting of: imaging, alignment, segmentation, ViT feature extraction, manifold projection.  }
    %3D RGB plots of the berry colors in a patch around a berry. The color plot illustrates the color variety of a single berry including lighter colors near the specularity. This plot depicts a patch around a cranberry and includes background leaf pixels. Segmentation of berries is used to isolate berry pixels, ideally excluding the contribution of leaf pixels. The observed  deep red color of the ripe cranberry indicates a high overheating risk.}
    \label{fig:ripenessMeasures}
    \vspace{-10pt}
\end{figure*}

The Mullica Queen variety has a relatively low risk of overheating based on its albedo class distribution for the first four collection dates. On the fifth collection date of 9/9, the number of red berries significantly increases, indicating that the berries' overheating risk is now high. This pattern is observed with slight variations over all three Mullica Queen cranberry beds. The Stevens variety has a majority of green berries for a significant portion of the collection period. However, by 9/9 it begins to cross over into the category for a high risk of overheating. 

The Crimson Queen variety crosses into the high risk category by 8/25. (Green albedo values  in the late-season graphs are an artifact due to mis-classification of some leaf pixels as berries.)
%We believe this to be due to an increase in false cranberry detections. 
The Haines variety crosses into the high risk category on 8/25 (in the sixth row) or on 8/31 (in the seventh row). 

From Figure \ref{fig:albedoOverTime}, we see that the Haines variety ripens the fastest. The next fastest ripening cranberry variety is Crimson Queen, followed by Mullica Queen. The Stevens variety is the slowest to ripen. These dates indicate a rough timeline indicating when cranberry farmers will need to monitor their crop more closely. These dates also serve as markers for when to focus more heavily on crop irrigation to mitigate overheating concerns.

\subsection{Individual Berry Ripening Metrics}

For individual berries, we follow the steps described in Sections~\ref{sec:individualImaging} and \ref{sec:individualBerryAnalysis}.
When comparing the four feature extractors shown in Figure~\ref{fig:cranvit}, we see that ViT features led to the most useful 2D
embedding showing ripeness progression in a well-distributed path close to a line. The other feature embeddings show a less clear progression from green to red, making them less easily interpretable. 
A ripeness metric is directly obtained by fitting the ViT-UMAP coordinates to a line and mapping initial values to zero ripeness and final values to unity. 
Plots of the ripeness values for the fourteen tracked individual berries are shown in Figure~\ref{fig:ripenessMeasures}. These plots show the  quantification of individual berry ripeness over time, and statistics of these ripeness metrics (e.g. mean and variance) over the bog show patterns for the cranberry variety. 

\section{Conclusion and Discussion}
%Touch upon: computer vision aid agriculture, 
%weakly supervised methods are important (point wise vs full-annotations),
% imaging, calibration and segmentation enables temporal albedo analysis, 
% the resulting temporal signatures give important predictive power to the growers enabling choices among cranberry crop varieties and implications of those choices in best agriculture practices (e.g. one size does not fit all for berries). 

We develop an Agtech framework for evaluating cranberry ripening patterns over multiple bogs using time-series imaging (both aerial and ground-based). 
We show the effectiveness of using foundation models for  combined with medium scale segmentation models tuned for cranberry segmentation. Classic computer vision methods for image registration (SIFT and homography estimation) are applied for  aligning the time-series images. Classic computer vision methods for basic radiometric/photometric calibration using gray-card measurements are also part of the workflow. 
%
\begin{comment}
Computer vision segmentation of photometrically calibrated images provides a real-time tool to assess crop health,  allow for expedient intervention, and compare crop varieties. We show the effectiveness of semantic segmentation on albedo characterization over time in cranberry crops. Weakly supervised semantic segmentation enables a convenient, effective localization  of cranberries  without the need for expensive pixel-wise labeling. We collect a time-series of weekly images from seven cranberry bogs over the course of two months using drones to 
create a labeled cranberry dataset with images throughout the entire cranberry growing cycle.
\end{comment}
%
This framework characterizes color development over time for cranberries and provides  key insight into berry overheating risk and crop health. We create a timeline of albedo change that gives farmers the tools to make more informed irrigation choices to prevent crop rot and conserve resources. The resulting temporal signatures give important predictive power to the growers enabling choices among cranberry crop varieties and implications of those choices in best agriculture practices. 
The methodology can be automated for large scale crop evaluation to support new methods of high throughput phenotyping. 
%(e.g. one size does not fit all for berries). 

% \section{Acknowledgments}
% This project was sponsored by the USDA NIFA AFRI, United States Award Number: 2019-67022-29922 and the SOCRATES NSF NRT \#2021628. We thank Michael King who supervised drone imagery collection for CRAID 2022.

\section{Acknowledgments}
This project was sponsored by the USDA NIFA AFRI Award Number: 2019-67022-29922 and the NSF NRT-FW-HTF: Socially Cognizant Robotics for a Technology Enhanced Society (SOCRATES) No. 2021628.

\newpage

\clearpage

%%%%%%%%% REFERENCES
{\small
\bibliographystyle{ieee_fullname}
\bibliography{egbib}
}

\end{document}